# Pseudo-Adaptive Penalization to Handle Constraints in Particle Swarm Optimizers

M. S. Innocente[1] and J. Sienz[1]
[1]ADOPT Research Group,
School of Engineering
Swansea University,
Swansea, UK

## Abstract

The penalization method is a popular technique to provide particle swarm optimizers with the ability to handle constraints. The downside is the need of penalization coefficients whose settings are problem-specific. While adaptive coefficients can be found in the literature, a different adaptive scheme is proposed in this paper, where coefficients are kept constant. A pseudo-adaptive relaxation of the tolerances for constraint violations while penalizing only violations beyond such tolerances results in a pseudo-adaptive penalization. A particle swarm optimizer is tested on a suite of benchmark problems for three types of tolerance relaxation: no relaxation; self-tuned initial relaxation with deterministic decrease; and self-tuned initial relaxation with pseudo-adaptive decrease. Other authors' results are offered as frames of reference.

## 1    Introduction

The particle swarm optimization paradigm is a population-based and gradient-free optimization method suitable for unconstrained problems. In order to be able to deal with constrained problems, some external mechanism needs to be incorporated. One of the most straightforward and popular techniques is the penalization method, where infeasible solutions are penalized by increasing their objective function value (minimization problems). The key issue is in the amount of penalization, which is typically linked to the amount of constraint violations. By turning the constrained problem into an unconstrained one, these methods are especially well suited for Particle Swarm Optimizers (PSOs) because they do not disrupt the normal dynamics of the swarm as, for instance, a repair algorithm would do. The drawback is that the penalization coefficients involved typically require problem-specific tuning. An excessive penalization might lead to premature convergence, whereas too mild a penalization might lead to infeasible solutions being chosen over feasible ones. Ideally, the penalization should be adaptive in terms of the problem and perhaps of the stage of the search, where initially low penalizations that are increased as the search progresses are advisable. A different adaptive scheme is proposed in this paper, where the penalization coefficients are kept constant. The procedure consists of an initially self-tuned relaxation of the constraint violation tolerance, followed by a pseudo-adaptive decrease of the relaxation. The self-tuning is performed so that an approximate target feasibility ratio is reached, which involves a number of constraint functions evaluations. The pseudo-adaptive decrease is linked to the number of potential feasible solutions found at the current time-step. Thus, by linking the penalization to the constraint violations beyond the pseudo-adaptive tolerance rather than to the actual constraint violations, a pseudo-adaptive penalization of infeasible solutions is achieved. For comparison, the same optimizer is tested using a deterministic exponential decrease of the tolerance relaxation, as well as without relaxing the desired tolerances.

The constrained optimization problem is posed in section 2; the particle swarm optimization paradigm is discussed in section 3; the proposed constraint-handling mechanism is presented in section 4; and the experimental results are offered in section 5. Conclusions and potential lines for future research are drawn in section 6.

## 2 Constrained optimization

Different families of optimization problems require different approaches to be dealt with. The first important difference is in the type of solution sought. Thus, problems whose optimum solution is a scalar or vector of scalars are called parameter optimization problems whereas problems whose optimum solution is a function or vector of functions are called variational problems. Another critical difference is in the type of variables, where a problem is referred to as continuous –or real-valued– optimization if the variables can take on real values and discrete optimization if they can only take on values from a given discrete set. Only real-valued, parameter optimization problems are considered within this paper.

The constraints bound the regions of the search-space where solutions are admissible. Thus, they may be due to geometric, technological, and/or physical restrictions, to name a few. While their linearity or nonlinearity does not make a difference in particle swarm optimization, their formulation as inequality or equality constraints does. The general, constrained, real-valued optimization problem may be formulated as follows:

$$\text{Minimize } f(\mathbf{x})$$
$$\text{subject to } \begin{cases} g_j(\mathbf{x}) \leq 0 & ; \quad j = 1, \ldots, q \\ g_j(\mathbf{x}) = 0 & ; \quad j = q+1, \ldots, q+r \\ l_i \leq x_i \leq u_i & ; \quad i = 1, \ldots, n \end{cases} \quad (1)$$

For a particle swarm optimizer to deal with equality constraints, a tolerance for their violations needs to be set, so that each equality constraint may be formulated in the form of a pair of inequality constraints. In turn, the latter can be expressed in the form of a single inequality constraint by means of the absolute function. Hence the problem is reformulated in terms of inequality constraints only as in Equation (2). Although the tolerance of inequality constraint violations is typically set to zero, it is set to the variable $Tol_{ineq}$ here both for generality and to allow its relaxation.

$$\text{Minimize } f(\mathbf{x})$$
$$\text{subject to } \begin{cases} g_j(\mathbf{x}) \leq Tol_{ineq} & ; \quad j = 1, \ldots, q \\ \text{abs}(g_j(\mathbf{x})) \leq Tol_{eq} & ; \quad j = q+1, \ldots, q+r \\ \max(0, x_i - u_i) + \max(0, -x_i + l_i) \leq 0 & ; \quad i = 1, \ldots, n \end{cases} \quad (2)$$

# 3  Particle swarm optimization

First introduced by James Kennedy and Russell C. Eberhart in 1995 [1], the particle swarm optimization (PSO) method has strong roots on different disciplines such as social psychology, artificial intelligence, and mathematical optimization. From the 'optimization' perspective, it is a gradient-free search method suitable for optimization problems whose solutions can be represented as points in an *n*-dimensional space. While variables need to be real-valued in its original version, binary and other discrete versions of the method have also been proposed. Refer, for instance, to [2, pp. 289-299]; [3]; [4]; and [5].

Since the method is not designed to optimize but to carry out procedures that are not directly related to the optimization problem, it is frequently referred to as a modern heuristics. Optimization occurs, nevertheless, without obvious links between the implemented technique and the resulting optimization process.

Influenced by Evolutionary Algorithms (EAs), the function to be minimized is commonly called 'fitness function'. It is more appropriately referred to as 'conflict function' hereafter due to the social-psychology metaphor that inspired the method.

Gradient information is not required, which enables the method to deal with non-differentiable and even discontinuous problems. Therefore there is no restriction to the characteristics of the objective function for the approach to be applicable. In fact, the function does not even need to be explicit.

The PSO algorithm is not designed to optimize but to perform a sort of simulation of a social milieu, where the ability of the population (swarm) to optimize its performance emerges from the cooperation among individuals (particles). While this makes it difficult to understand how optimization is actually performed, it shows astonishing robustness in handling different kinds of complex problems that it was not specifically designed for. It has the disadvantage that its theoretical bases are very difficult to be understood deterministically. Nevertheless, considerable theoretical work has been carried out on simplified versions of the algorithm (e.g. [6], [7], [2], [8], [9], [10], and [11]. For a comprehensive review of the method, refer to [12] and [13].

## 3.1 Basic algorithm

While the emergent optimization properties of the PSO algorithm result from local interactions among particles in a swarm, the behaviour of a single particle can be summarized in three sequential processes: *evaluation*; *comparison*; and *imitation*. Thus, the performance of a particle in its current position is evaluated in terms of the conflict function. In order to decide upon its next position, the particle compares its current conflict to those associated with its own and with its neighbours' best experiences. Finally, the particle imitates its own best experience and the best experience in its neighbourhood to some extent. The basic update equations are:

$$\begin{cases} v_{ij}^{(t)} = w \cdot v_{ij}^{(t-1)} + iw \cdot U_{(0,1)} \cdot \left(pbest_{ij}^{(t-1)} - x_{ij}^{(t-1)}\right) + sw \cdot U_{(0,1)} \cdot \left(lbest_{ij}^{(t-1)} - x_{ij}^{(t-1)}\right) \\ x_{ij}^{(t)} = x_{ij}^{(t-1)} + v_{ij}^{(t)} \end{cases} \quad (3)$$

where:

$v_{ij}^{(t)}$ : $j^{th}$ coordinate of the velocity of particle *i* at time-step *t*.

$x_{ij}^{(t)}$ : $j^{th}$ coordinate of the position of particle *i* at time-step *t*.

$U_{(0,1)}$ : Random number from a uniform distribution in the range [0,1] resampled anew every time it is referenced.

*w*, *iw*, *sw* : Inertia, individuality, and sociality weights, respectively.

$pbest_{ij}^{(t)}$ : $j^{th}$ coordinate of the best position found by particle *i* by time-step *t*.

$lbest_{ij}^{(t)}$ : $j^{th}$ coordinate of the best position found by any particle in the neighbourhood of particle *i* by time-step *t*.

As it can be observed, there are three coefficients that govern the dynamics of the swarm: the inertia (*w*), the individuality (*iw*), and the sociality (*sw*) weights. The settings of these coefficients greatly affect the behaviour of the swarm. Equation (3) can also be formulated in terms of $\phi_i$ and $\phi_s$ as shown in Equation (4):

$$\begin{cases} v_{ij}^{(t)} = w \cdot v_{ij}^{(t-1)} + \phi_i \cdot \left(pbest_{ij}^{(t-1)} - x_{ij}^{(t-1)}\right) + \phi_s \cdot \left(lbest_{ij}^{(t-1)} - x_{ij}^{(t-1)}\right) \\ 0 \leq \left(\phi_i + \phi_s = iw \cdot U_{(0,1)} + sw \cdot U_{(0,1)}\right) \leq \left(aw = iw + sw\right) \\ x_{ij}^{(t)} = x_{ij}^{(t-1)} + v_{ij}^{(t)} \end{cases} \quad (4)$$

## 3.2 Two alternative formulations

Two alternative formulations were proposed by Innocente [11], consisting of defining a desired average behaviour and then adding randomness as a type of noise. Thus the strength of the attraction towards the weighted average of the best experiences is no longer a random value between zero and the acceleration weight. Instead the random value is within the range $[\phi_{min}, \phi_{max}]$, and the acceleration weight is the centre of the interval rather than its upper limit. Furthermore, the individuality and sociality weights are replaced by coefficients $ip \in [0,1)$ and $sp \in (0,1]$, where their aggregation $ip + sp = 1$. The procedure starts with the user selecting the desired strength awarded to the individuality and the sociality by choosing $ip \in [0,1)$. The common formulation for both proposed alternatives is shown in Equation (5).

$$\begin{cases} v_{ij}^{(t)} = w \cdot v_{ij}^{(t-1)} + \phi_i \cdot \left(pbest_{ij}^{(t-1)} - x_{ij}^{(t-1)}\right) + \phi_s \cdot \left(lbest_{ij}^{(t-1)} - x_{ij}^{(t-1)}\right) \\ \phi_i = ip \cdot \left[\phi_{min} + (\phi_{max} - \phi_{min}) \cdot U_{(0,1)}\right] \\ \phi_s = sp \cdot \left[\phi_{min} + (\phi_{max} - \phi_{min}) \cdot U_{(0,1)}\right] \\ ip \in [0,1) \quad ; \quad sp = 1 - ip \\ x_{ij}^{(t)} = x_{ij}^{(t-1)} + v_{ij}^{(t)} \end{cases} \quad (5)$$

**PSO-RRR1**: The first alternative consists of selecting $aw \in (1.00, 2.00)$, and

$$\begin{cases} w = aw - 1 \\ \phi_{max} = \frac{3}{2} \cdot (w+1) \\ \phi_{min} = \frac{1}{2} \cdot (w+1) \end{cases} \quad (6)$$

**PSO-RRR2**: The second alternative consists of selecting $aw \in (1.00, 2.61]$, and

$$\begin{cases} w = \frac{1}{aw} - 2 + aw \\ \phi_{max} = 2 \cdot (w+1) \\ \phi_{min} = 2 \cdot aw - \phi_{max} \end{cases} \quad (7)$$

It is advisable not to take values of *aw* too close to the limits. For further details on these two formulations, refer to [11], chapter 6.

## 4 Pseudo-adaptive penalization

### 4.1 Static penalization

Penalization methods can be viewed as optimizing two objectives: minimizing the objective function on one hand and minimizing constraint violations on the other, where the second objective is already an aggregation of objectives (minimizing each constraint function violation). These methods combine all these objectives into a single function to be optimized, thus turning the constrained problem into an unconstrained one where the relative priority awarded to the different objectives is somewhat weighted. The main concept is that this new function must coincide with the original one when every constraint is satisfied. Hence the objective function is penalized for infeasible solutions only. Different kinds of penalization methods can be found in the literature according to the way the penalization is calculated.

The advantage of these methods is that they use both objective and constraint functions information within the infeasible space to smoothly guide the search towards more promising areas. Since constrained problems are treated as unconstrained ones once infeasible solutions are penalized, they work well on highly constrained problems. The drawback is that they are sensitive to the tuning of at least a couple of problem-dependent penalization coefficients. High penalizations might lead to infeasible regions not being explored converging to non-optimal but feasible solutions, whereas low penalizations might lead to the system evolving solutions that are violating constraints but present themselves as better than feasible solutions. However, research on adaptive coefficients is extensive in the literature (e.g. [14] and [15]). A classical additive, constant penalization scheme –linked to the amount of infeasibility– is shown in Equations (8) and (9):

$$f_p(\mathbf{x}) = f(\mathbf{x}) + \sum_{j=1}^{m} \left[ k_j \cdot \left(f_j(\mathbf{x})\right)^{\alpha_j} \right] \qquad (8)$$

$$f_j(\mathbf{x}) = \begin{cases} \max\{0, g_j(\mathbf{x})\} & ; \quad 1 \leq j \leq q \\ \mathrm{abs}(g_j(\mathbf{x})) & ; \quad q < j \leq m \end{cases} \qquad (9)$$

where $f(\mathbf{x})$ is the conflict function; $f_p(\mathbf{x})$ is the penalized conflict function; $f_j(\mathbf{x})$ is the amount of violation of $j^{\text{th}}$ constraint; and $k_j$ and $\alpha_j$ are penalization coefficients. These coefficients may be constant, time-varying or adaptive, and they can be the same or different for different constraints. Typically, $k_j$ is set to high and $\alpha_j$ to small values. It is not recommendable to use different penalization coefficients for different constraints because that makes the coefficients' tuning more difficult for every problem. An alternative to account for the different sensitivity of the penalized objective function to the different constraint violations which may be of different orders of magnitude consists of normalizing the constraint violations. Even further, the original conflict function may also be normalized so that the latter and the overall measure of constraint violations are of the same order of magnitude. These normalizations are not considered here and therefore left for future work.

## 4.2 Proposed pseudo-adaptive penalization

As previously mentioned, arbitrarily set –not tuned– constant coefficients are used in the experiments hereafter. The study of their convenient setting is left for future research, where the hope is that the normalization of constraints and perhaps of the conflict function would allow for their removal from Equation (8). Thus, the penalization scheme proposed in this paper is as shown in Equations (10) to (12), where the adaptiveness lies in the (pseudo) adaptiveness of the tolerances.

$$f_p(\mathbf{x}) = f(\mathbf{x}) + k \cdot \sum_{j=1}^{m} (f_j(\mathbf{x}))^\alpha \qquad (10)$$

$$f_j(\mathbf{x}) = \begin{cases} \max\{0, (g_j(\mathbf{x}) - Tol_{ineq})\} & ; \quad 1 \le j \le q \\ \max\{0, (\text{abs}(g_j(\mathbf{x})) - Tol_{eq})\} & ; \quad q < j \le m \end{cases} \qquad (11)$$

$$k = 10^6$$
$$\alpha = \begin{cases} 2 & \text{if } f_j(\mathbf{x}) \ge 1 \\ 1 & \text{if } f_j(\mathbf{x}) < 1 \end{cases} \qquad (12)$$

### 4.2.1 Self-tuned initial tolerances relaxation

The use of tolerances for equality constraint violations in population-based methods is of common practice. It is also not uncommon to relax such tolerances, where the decrease of such relaxations is typically deterministic. The aim of these relaxations is to temporarily increase the feasible region of the search-space. However, the impact of a given relaxation on the feasibility ratio (FR) of the search-space is problem-dependent, and can vary greatly. For instance, to obtain a $FR \in [20\%, 25\%]$, a tolerance for equality constraint violations of around 0.26 is required for the problem g11, whereas a tolerance of around 6.63 is required for problem g13 (both problems involving equality constraints only). In addition, since there are problems involving only inequality constraints that present very small FRs, the same concept can be applied. That is, the tolerance for inequality constraint violations can also be dynamically relaxed. Note, for instance, that the tolerance required for problem g10 to present a $FR \in [20\%, 25\%]$ is around 10.83, whereas it is around 2790 for problem g06 (both involving inequality constraints only). These examples clearly illustrate how problem-dependent the effect of a given tolerance may be.

Hence an initial self-tuned tolerance relaxation is proposed aiming for a desired FR. Thus, the self-tuning procedure consists of starting with a small, minimum value for the tolerance and evaluating the constraint functions of 1000 randomly selected solutions. The FR is evaluated, and the tolerance is adequately increased or decreased. For problems involving inequality and equality constraints, the tolerance for the violations of equality constraints are arbitrarily kept 10 times greater than that of inequality constraint violations. For further details on the implementation of this procedure, refer to [11], chapter 8.

### 4.2.2 Pseudo-adaptive decrease of tolerances relaxation

The aim is to make the tolerance update adaptive so that updates are performed when they would have a less disruptive effect in maintaining potentially good solutions. Hence updates are performed when a given percentage of the particles' best experiences (PBESTs) are located within feasible space.

The coefficient for the exponential update is also pseudo-adaptive as shown in Equation (13), where $per^{(t)}$ is the current percentage of feasible PBESTs. The exponential update of the tolerances is as posed in Equation (14). Thus $ktol^{(t)} = 0.99$ for $per^{(t)} = per_{min}$; $ktol^{(t)} = ktol_{min}$ for $per^{(t)} = 100$; and the variation in between is linear. Therefore the greater the percentage above a minimum established the smaller the value of $ktol^{(t)}$, and hence the greater the tolerance decrease. Obviously, there is no update for $per^{(t)} < per_{min}$.

$$ktol^{(t)} = \frac{0.99 - ktol_{min}}{100 - per_{min}} \cdot \left(100 - per^{(t)}\right) + ktol_{min} \qquad (13)$$

$$Tol^{(t)} = ktol^{(t)} \cdot Tol^{(t-1)} \qquad (14)$$

Since the tolerance for inequality constraint violations is typically set to zero, whenever it goes below $10^{-5}$ it is automatically reset to zero.

Aiming to avoid too many time-steps without a tolerance update, a safety mechanism is implemented by enforcing a tolerance update if

$$\frac{t}{n^o \ tol. \ updates} \geq 20 \qquad (15)$$

where $t$ is the number of time-step and $n^o$ *tol. updates* is the number of tolerance updates performed so far. When the update is enforced by Equation (15), the coefficient used in Equation (14) equals $ktol^{(t)} = 0.99$.

In order to give some time for the particles to find feasible solutions once the tolerances have reached their desired value, it is arbitrarily set that such values are reached by the time 80% of the search has been carried out ($t_{min}$). If the desired tolerance was not reached by $t = 0.9 \cdot t_{min}$, an update is performed at every time-step (i.e. from $t = 0.9 \cdot t_{min} + 1$ to $t = t_{min}$) using the coefficient in Equation (16). Thus final tolerances are ensured to be reached by $t = t_{min}$.

$$ktol = \left(\frac{Tol^{(t_{min})}}{Tol^{(0.9 \cdot t_{min})}}\right)^{\frac{1}{0.1 \cdot t_{min}}} \qquad (16)$$

where *ktol* is calculated independently for inequality and equality constraints. Since $Tol_{ineq}^{(t_{min})} = 0$, the latter is replaced by $10^{-5}$ for the calculation of *ktol* in Equation (16) and $Tol_{ineq}$ is set to zero as soon as it reaches a value below or equal to $10^{-5}$. Of course $t_{min}$, $0.9 \cdot t_{min}$, and $0.1 \cdot t_{min}$ are rounded-off to integer values if necessary.

# 5 Experimental results

The details of the implementation for the experiments are as follows: 50 particles; 10000 time-steps, constant penalization coefficients $k = 10^6$ and $\alpha = 2$ in Equation (10); the best of 1000 Latin Hypercube samplings for the particles' initialization according to the maximum minimum distance criterion; velocities initialized to zero while the best individual experiences are initialized in the same manner as the initial positions; synchronous update of the best experiences; the forward topology with three sub-neighbourhoods as proposed by Innocente [11]; the coefficients and formulations for the first sub-neighbourhood are those of the PSO-RRR2 with $aw = 2.40$ as in Equations (5) and (7); the coefficients and formulations for the second sub-neighbourhood are those of the PSO-RRR1 with $aw = 1.80$ as in Equations (5) and (6); the coefficients and formulations for the third sub-neighbourhood are those of the classical PSO with $w = 0.7298$ and $aw = 2.9922$ as in Equation (3) –also coinciding with Clerc et al.'s constriction factor *type 1"* [9] with $cf = 0.7298$ and $aw = 4.10$–; initial self-tuned tolerance relaxation such that $FR = 20-25\%$. If the FR of the problem is already greater than that, it is increased in around 5%. If there are inequality and equality constraints, the tolerance for the violations of the latter are arbitrarily set 10 times greater than that of the inequality constraints. The results obtained with no relaxation are also offered for comparison. For relaxed tolerances, two types of decrease of the initial ones are considered: the proposed pseudo-adaptive scheme, and an exponential decrease with a constant coefficient $ktol^{(t)} = 0.98$. For the pseudo-adaptive decrease, $ktol_{min} = 0.90$ and $per_{min} = 80\%$ in Equation (13), while $t_{min} = 0.80 \cdot t_{max} = 8000$ in Equation (16).

The proposed pseudo-adaptive penalization method is tested on a well known benchmark suite composed of 13 constrained problems. Their main features are offered in Table 1 together with their approximate FRs without tolerances, with final tolerances, and with mean initial tolerances. 25 runs are performed for the statistics. The values of the mean initial tolerances that result in the desired initial FRs are also provided in the table.

The results obtained for the three different types of tolerance relaxations are presented in Table 2, including the exact number of objective and constraint function evaluations. A solution is considered successful if the error is not greater than $10^{-4}$. The 'mean [%] feasible PBESTs' is the mean percentage –among 25 runs– of individual best experiences of the particles that are feasible in the final time-step.

For comparison, the results obtained for the same suite of benchmark problems by two other PSO algorithms in the literature –namely those proposed by Toscano Pulido et al. [16] and by Muñoz Zavala et al. [17]– are offered in Table 3 together with the results obtained by the pseudo-adaptive approach proposed here.

Since the values of the tolerances are pseudo-adaptive, it is interesting to observe the form of the curves of their evolution throughout the search. Due to the stochastic nature of the paradigm, those curves vary from one run to the next for a given problem. The curves showing the evolution of the tolerances corresponding to four selected problems are offered in Figure 1 to Figure 4. Thus, Figure 1 corresponds to problem g01 involving 9 inequality constraints; Figure 2 corresponds to problem

g03 involving 1 equality constraint only; Figure 3 corresponds to problem g05 involving 3 equality constraints and 1 inequality constraint (in the form of an interval); and Figure 4 corresponds to problem g13 involving 3 equality constraints. In those figures, the plain tolerance curves correspond to one single arbitrary run whereas the average curves correspond to the average tolerances among all 25 runs.

| Problem | Optimum | Dim. | IC | EC | FR [%] | FR [%] for desired tolerance | FR [%] for initial tolerance | Mean initial inequality tolerance | Mean initial equality tolerance |
|---|---|---|---|---|---|---|---|---|---|
| g01 | -15.000000 | 13 | 9 | 0 | 0.0003 | 0.0003 | 23.4617 | 89.92 | N/A |
| g02 | -0.803619 | 20 | 2 | 0 | 99.9971 | 99.9971 | 99.9971 | 0.01 | N/A |
| g03 | -1.000500 | 10 | 0 | 1 | < 0.0001 | 0.0002 | 24.5335 | N/A | 1.66 |
| g04 | -30665.538672 | 5 | 3 (#) | 0 | 26.9887 | 26.9887 | 30.2026 | 0.11 | N/A |
| g05 | 5126.496714 | 4 | 1 (*) | 3 | < 0.0001 | < 0.0001 | 23.3053 | 68.88 | 688.79 |
| g06 | -6961.813876 | 2 | 2 | 0 | 0.0074 | 0.0074 | 24.3050 | 2790.51 | N/A |
| g07 | 24.306209 | 10 | 8 | 0 | 0.0001 | 0.0001 | 23.8399 | 383.89 | N/A |
| g08 | -0.095825 | 2 | 2 | 0 | 0.8610 | 0.8610 | 23.4371 | 9.88 | N/A |
| g09 | 680.630057 | 7 | 4 | 0 | 0.5232 | 0.5232 | 24.0533 | 421.13 | N/A |
| g10 | 7049.248021 | 8 | 6 | 0 | 0.0005 | 0.0005 | 21.1715 | 10.83 | N/A |
| g11 | 0.749900 | 2 | 0 | 1 | <0.0001 | 0.0108 | 24.8914 | N/A | 0.26 |
| g12 | -1.000000 | 3 | 1 (@) | 0 | 4.7713 | 4.7713 | 22.0256 | 0.11 | N/A |
| g13 | 0.053942 | 5 | 0 | 3 | <0.0001 | <0.0001 | 22.8845 | N/A | 6.63 |

(#) Other authors claim there are 6 inequality constraints, but each one defines an interval, therefore no more than 3 constraints can be violated simultaneously.

(*) Other authors claim there are two inequality constraints, but they define an interval, so that no more than 1 constraint can be violated simultaneously.

(@) Most authors claim there are $9^3$ inequality constraints, but in reality it is one constraint that splits the feasible space in $9^3$ (729) disjointed spheres. The solution needs to be inside one sphere to be feasible, so that membership to all 729 spheres is not possible. If the constraint is viewed as 729 constraints, then at least 728 of them will be always violated.

Table 1. Features of the problems in the test suite: number of dimensions, inequality and equality constraints; feasibility ratios (FRs) of the problem with no tolerance, desired tolerance, and initial tolerance; and the mean self-tuned initial inequality and equality tolerances. FRs are calculated by randomly generating $10^6$ solutions, where final (desired) equality constraint violations tolerance equals $10^{-4}$

Due to the tolerance relaxations, intermediate solutions that are temporarily regarded as feasible smaller than the actual feasible minimum solution of the problem can be found by the optimizer. In these cases, the best solution might increase rather than decrease as the search progresses. An example of this is shown in Figure 5 for problem g05.

# 6 Conclusions and future research

The results in Table 2 show that the pseudo-adaptive scheme obtains the best results overall, in terms of the best, median and mean solutions found. The technique results in remarkable improvement compared to the same optimizer without relaxation, especially in problems with equality constraints such as g03, g05, g11 and g13. The exception in this test suite is in problem g10, for which the solution is decreased by the pseudo-adaptive scheme. The reason for this is still to be studied.

| Problem | Optimum | Type of tolerance relaxation | BEST | MEDIAN | MEAN | WORST | [%] Feasible Solutions | [%] Successful Solutions | Mean FEs | Mean CEs | Mean [%] Feasible PBESTs |
|---|---|---|---|---|---|---|---|---|---|---|---|
| g01 | -15.000000 | NONE | -15.000000 | -15.000000 | -15.000000 | -15.000000 | 100.00 | 100.00 | 5.00E+05 | 5.00E+05 | 99.92 |
| | | EXP. | -15.000000 | -15.000000 | -15.000000 | -15.000000 | 100.00 | 100.00 | 5.00E+05 | 5.86E+05 | 99.76 |
| | | ADAPTIVE | -15.000000 | -15.000000 | -15.000000 | -15.000000 | 100.00 | 100.00 | 5.00E+05 | 5.75E+05 | 99.36 |
| g02 | -0.803619 | NONE | -0.803618 | -0.794896 | -0.792568 | -0.687854 | 100.00 | 40.00 | 5.00E+05 | 5.00E+05 | 100.00 |
| | | EXP. | -0.803618 | -0.803429 | -0.794852 | -0.758093 | 100.00 | 48.00 | 5.00E+05 | 5.35E+05 | 100.00 |
| | | ADAPTIVE | -0.803618 | -0.803429 | -0.794852 | -0.758093 | 100.00 | 48.00 | 5.00E+05 | 5.08E+05 | 100.00 |
| g03 | -1.000500 | NONE | -0.999213 | -0.983214 | -0.972554 | -0.894905 | 100.00 | 0.00 | 5.00E+05 | 5.00E+05 | 98.80 |
| | | EXP. | -1.000493 | -1.000477 | -1.000467 | -1.000326 | 100.00 | 96.00 | 5.00E+05 | 5.77E+05 | 100.00 |
| | | ADAPTIVE | -1.000499 | -1.000496 | -1.000493 | -1.000459 | 100.00 | 100.00 | 5.00E+05 | 5.84E+05 | 99.92 |
| g04 | -30665.538672 | NONE | -30665.538672 | -30665.538672 | -30665.538672 | -30665.538672 | 100.00 | 100.00 | 5.00E+05 | 5.00E+05 | 100.00 |
| | | EXP. | -30665.538672 | -30665.538672 | -30665.538672 | -30665.538672 | 100.00 | 100.00 | 5.00E+05 | 5.50E+05 | 100.00 |
| | | ADAPTIVE | -30665.538672 | -30665.538672 | -30665.538672 | -30665.538672 | 100.00 | 100.00 | 5.00E+05 | 5.14E+05 | 100.00 |
| g05 | 5126.496714 | NONE | 5126.498381 | 5158.465976 | 5242.672049 | 5708.280940 | 100.00 | 0.00 | 5.00E+05 | 5.00E+05 | 25.20 |
| | | EXP. | 5126.516036 | 5155.072539 | 5235.566090 | 5885.912510 | 100.00 | 0.00 | 5.00E+05 | 5.87E+05 | 24.80 |
| | | ADAPTIVE | 5126.593807 | 5130.122719 | 5142.265330 | 5318.299822 | 100.00 | 0.00 | 5.00E+05 | 6.31E+05 | 57.52 |
| g06 | -6961.813876 | NONE | -6961.813876 | 6961.813876 | -6961.813876 | -6961.813876 | 100.00 | 100.00 | 5.00E+05 | 5.00E+05 | 100.00 |
| | | EXP. | -6961.813876 | -6961.813876 | -6961.813876 | -6961.813876 | 100.00 | 100.00 | 5.00E+05 | 6.18E+05 | 100.00 |
| | | ADAPTIVE | -6961.813876 | -6961.813876 | -6961.813876 | -6961.813876 | 100.00 | 100.00 | 5.00E+05 | 6.38E+05 | 100.00 |
| g07 | 24.306209 | NONE | 24.308588 | 24.397361 | 24.447352 | 25.185085 | 100.00 | 0.00 | 5.00E+05 | 5.00E+05 | 100.00 |
| | | EXP. | 24.322332 | 24.438153 | 24.447324 | 24.796222 | 100.00 | 0.00 | 5.00E+05 | 6.03E+05 | 100.00 |
| | | ADAPTIVE | 24.322181 | 24.483518 | 24.515330 | 24.948167 | 100.00 | 0.00 | 5.00E+05 | 5.58E+05 | 100.00 |
| g08 | -0.095825 | NONE | -0.095825 | -0.095825 | -0.095825 | -0.095825 | 100.00 | 100.00 | 5.00E+05 | 5.00E+05 | 100.00 |
| | | EXP. | -0.095825 | -0.095825 | -0.095825 | -0.095825 | 100.00 | 100.00 | 5.00E+05 | 5.72E+05 | 100.00 |
| | | ADAPTIVE | -0.095825 | -0.095825 | -0.095825 | -0.095825 | 100.00 | 100.00 | 5.00E+05 | 5.20E+05 | 100.00 |
| g09 | 680.630057 | NONE | 680.630477 | 680.632574 | 680.633039 | 680.638835 | 100.00 | 0.00 | 5.00E+05 | 5.00E+05 | 100.00 |
| | | EXP. | 680.630753 | 680.632033 | 680.632550 | 680.636595 | 100.00 | 0.00 | 5.00E+05 | 6.03E+05 | 100.00 |
| | | ADAPTIVE | 680.630093 | 680.632446 | 680.632900 | 680.637958 | 100.00 | 8.00 | 5.00E+05 | 5.36E+05 | 100.00 |
| g10 | 7049.248021 | NONE | 7059.928996 | 7154.036097 | 7169.147324 | 7440.039736 | 100.00 | 0.00 | 5.00E+05 | 5.00E+05 | 98.72 |
| | | EXP. | 7049.728978 | 7105.765355 | 7140.793308 | 7348.447274 | 100.00 | 0.00 | 5.00E+05 | 5.73E+05 | 97.04 |
| | | ADAPTIVE | 7118.872990 | 7489.944970 | 7570.781098 | 8155.654975 | 96.00 | 0.00 | 5.00E+05 | 6.20E+05 | 82.88 |
| g11 | 0.749900 | NONE | 0.749900 | 0.749900 | 0.749901 | 0.749915 | 100.00 | 100.00 | 5.00E+05 | 5.00E+05 | 99.20 |
| | | EXP. | 0.749900 | 0.749900 | 0.749907 | 0.749975 | 100.00 | 100.00 | 5.00E+05 | 5.19E+05 | 99.68 |
| | | ADAPTIVE | 0.749900 | 0.749900 | 0.749900 | 0.749903 | 100.00 | 100.00 | 5.00E+05 | 5.95E+05 | 90.24 |
| g12 | -1.000000 | NONE | -1.000000 | -1.000000 | -1.000000 | -1.000000 | 100.00 | 100.00 | 5.00E+05 | 5.00E+05 | 100.00 |
| | | EXP. | -1.000000 | -1.000000 | -1.000000 | -1.000000 | 100.00 | 100.00 | 5.00E+05 | 5.48E+05 | 100.00 |
| | | ADAPTIVE | -1.000000 | -1.000000 | -1.000000 | -1.000000 | 100.00 | 100.00 | 5.00E+05 | 5.16E+05 | 100.00 |
| g13 | 0.053942 | NONE | 0.170131 | 0.600263 | 0.632353 | 0.983725 | 100.00 | 0.00 | 5.00E+05 | 5.00E+05 | 38.80 |
| | | EXP. | 0.277980 | 0.720584 | 0.706277 | 0.996941 | 100.00 | 0.00 | 5.00E+05 | 5.08E+05 | 40.96 |
| | | ADAPTIVE | 0.053943 | 0.054119 | 0.131239 | 0.439679 | 100.00 | 36.00 | 5.00E+05 | 6.29E+05 | 79.92 |

Table 2. Statistical results obtained for the 13 problems in the test suite for three types of tolerance relaxation: none, initially self-tuned with exponential decrease, and initially self-tuned with (pseudo) adaptive decrease. The percentages of feasible solutions, successful solutions (error not greater than $10^{-4}$), the mean numbers of FEs and CEs, and the mean percentage of feasible PBESTs at the end of the search are also provided.

| Problem | Optimum | Type of tolerance relaxation | BEST | MEDIAN | MEAN | WORST | [%] Feasible Solutions | [%] Successful Solutions | Mean FEs | Mean CEs | Mean [%] Feasible PBESTs |
|---|---|---|---|---|---|---|---|---|---|---|---|
| g01 | -15.000000 | ADAPTIVE | -15.000000 | -15.000000 | -15.000000 | -15.000000 | 100.00 | 100.00 | 5.00E+05 | 5.75E+05 | 25 |
|  |  | Toscano | -15.000000 | - | -15.000000 | -15.000000 | - | - | 3.40E+05 | - | 30 |
|  |  | PESO | -15.000000 | -15.000000 | -15.000000 | -15.000000 | - | - | 3.40E+05 | - | 30 |
| g02 | -0.803619 | ADAPTIVE | -0.803618 | -0.803429 | -0.794852 | -0.758093 | 100.00 | 48.00 | 5.00E+05 | 5.08E+05 | 25 |
|  |  | Toscano | -0.803432 | - | -0.790400 | -0.750393 | - | - | 3.40E+05 | - | 30 |
|  |  | PESO | -0.792608 | -0.731693 | -0.721749 | 0.614135 | - | - | 3.40E+05 | - | 30 |
| g03 | -1.000500 | ADAPTIVE | -1.000499 | -1.000496 | -1.000493 | -1.000459 | 100.00 | 100.00 | 5.00E+05 | 5.84E+05 | 25 |
|  |  | Toscano | -1.004720 | - | -1.003814 | -1.002490 | - | - | 3.40E+05 | - | 30 |
|  |  | PESO | -1.005010 | -1.005008 | -1.005006 | -1.004989 | - | - | 3.40E+05 | - | 30 |
| g04 | -30665.538672 | ADAPTIVE | -30665.538672 | -30665.538672 | -30665.538672 | -30665.538672 | 100.00 | 100.00 | 5.00E+05 | 5.14E+05 | 25 |
|  |  | Toscano | -30665.500000 | -30665.500000 | -30665.500000 | -30665.500000 | - | - | 3.40E+05 | - | 30 |
|  |  | PESO | -30665.538672 | -30665.538672 | -30665.538672 | -30665.538672 | - | - | 3.40E+05 | - | 30 |
| g05 | 5126.496714 | ADAPTIVE | 5126.593807 | 5130.122719 | 5142.265330 | 5318.299822 | 100.00 | 0.00 | 5.00E+05 | 6.31E+05 | 25 |
|  |  | Toscano | 5126.640000 | - | 5461.081333 | 6104.750000 | - | - | 3.40E+05 | - | 30 |
|  |  | PESO | 5126.484154 | 5126.538302 | 5129.178298 | 5148.859414 | - | - | 3.40E+05 | - | 30 |
| g06 | -6961.813876 | ADAPTIVE | -6961.813876 | -6961.813876 | -6961.813876 | -6961.813876 | 100.00 | 100.00 | 5.00E+05 | 6.38E+05 | 25 |
|  |  | Toscano | -6961.810000 | - | -6961.810000 | -6961.810000 | - | - | 3.40E+05 | - | 30 |
|  |  | PESO | -6961.813876 | -6961.813876 | -6961.813876 | -6961.813876 | - | - | 3.40E+05 | - | 30 |
| g07 | 24.306209 | ADAPTIVE | 24.322181 | 24.483518 | 24.515330 | 24.948167 | 100.00 | 0.00 | 5.00E+05 | 5.58E+05 | 25 |
|  |  | Toscano | 24.351100 | - | 25.355771 | 27.316800 | - | - | 3.40E+05 | - | 30 |
|  |  | PESO | 24.306921 | 24.371253 | 24.371253 | 24.593504 | - | - | 3.40E+05 | - | 30 |
| g08 | -0.095825 | ADAPTIVE | -0.095825 | -0.095825 | -0.095825 | -0.095825 | 100.00 | 100.00 | 5.00E+05 | 5.20E+05 | 25 |
|  |  | Toscano | -0.095825 | - | -0.095825 | -0.095825 | - | - | 3.40E+05 | - | 30 |
|  |  | PESO | -0.095825 | -0.095825 | 0.095825 | -0.095825 | - | - | 3.40E+05 | - | 30 |
| g09 | 680.630057 | ADAPTIVE | 680.630093 | 680.632446 | 680.632900 | 680.637958 | 100.00 | 8.00 | 5.00E+05 | 5.36E+05 | 25 |
|  |  | Toscano | 680.638000 | - | 680.852393 | 681.553000 | - | - | 3.40E+05 | - | 30 |
|  |  | PESO | 680.630057 | 680.630057 | 680.630057 | 680.630058 | - | - | 3.40E+05 | - | 30 |
| g10 | 7049.248021 | ADAPTIVE | 7118.872990 | 7489.944970 | 7570.781098 | 8155.654975 | 96.00 | 0.00 | 5.00E+05 | 6.20E+05 | 25 |
|  |  | Toscano | 7057.590000 | - | 7560.047857 | 8104.310000 | - | - | 3.40E+05 | - | 30 |
|  |  | PESO | 7049.459452 | 7069.926219 | 7099.101385 | 7251.396244 | - | - | 3.40E+05 | - | 30 |
| g11 | 0.749900 | ADAPTIVE | 0.749900 | 0.749900 | 0.749900 | 0.749903 | 100.00 | 100.00 | 5.00E+05 | 5.95E+05 | 25 |
|  |  | Toscano | 0.749999 | - | 0.750107 | 0.752885 | - | - | 3.40E+05 | - | 30 |
|  |  | PESO | 0.749000 | 0.749000 | 0.749000 | 0.749000 | - | - | 3.40E+05 | - | 30 |
| g12 | -1.000000 | ADAPTIVE | -1.000000 | -1.000000 | -1.000000 | -1.000000 | 100.00 | 100.00 | 5.00E+05 | 5.16E+05 | 25 |
|  |  | Toscano | -1.000000 | - | -1.000000 | -1.000000 | - | - | 3.40E+05 | - | 30 |
|  |  | PESO | -1.000000 | -1.000000 | -1.000000 | -1.000000 | - | - | 3.40E+05 | - | 30 |
| g13 | 0.053942 | ADAPTIVE | 0.053943 | 0.054119 | 0.131239 | 0.439679 | 100.00 | 36.00 | 5.00E+05 | 6.29E+05 | 25 |
|  |  | Toscano | 0.068665 | - | 1.716426 | 13.669500 | - | - | 3.40E+05 | - | 30 |
|  |  | PESO | 0.081498 | 0.631946 | 0.626881 | 0.997586 | - | - | 3.40E+05 | - | 30 |

Table 3. Statistical results obtained for the 13 problems in the test suite for the (pseudo) adaptive penalization scheme proposed, together with those reported by Toscano Pulido et al. [16] and by Muñoz Zavala et al. [17] as references.

It is quite surprising to observe the improvement in the quality of the solutions resulting from the pseudo-adaptive scheme on problem g02, which presents a very high FR. The reason for this is probably that the solution lies near the boundary of feasible space, and the relaxation of the tolerance allows particles to approach the solution from every direction rather than from feasible space only.

The self-tuned initial relaxation with deterministic, exponential decrease presents competitive or better performance than no relaxation on all problems in terms of the best, median, and mean solutions found, except for g13. The surprising decrease in the quality of the solutions for problem g13 –while the pseudo-adaptive decrease leads to remarkable improvement– seems to suggest that updating the tolerance 'too soon' may turn the relaxation of tolerances into a harmful mechanism, as potentially good solutions may be lost during the updates.

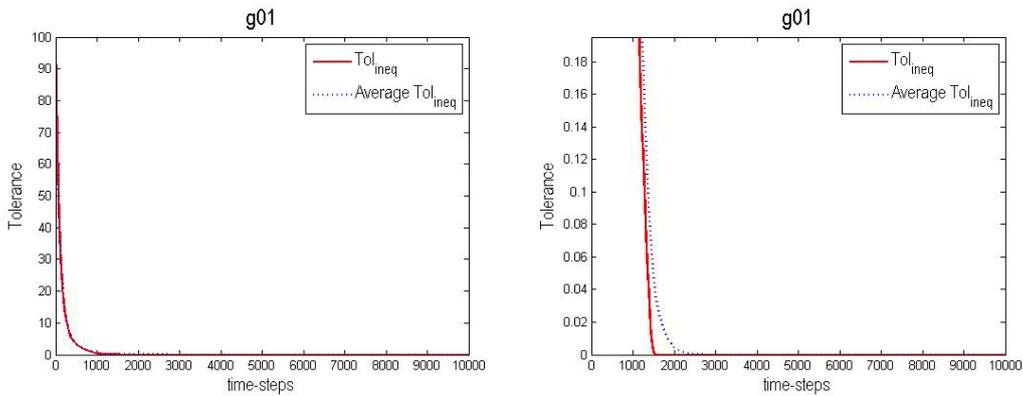

Figure 1: Pseudo-adaptive tolerance for inequality constraint violations in problem g01. The average is among the 25 runs. The figure on the right is just a zoom of the one on the left.

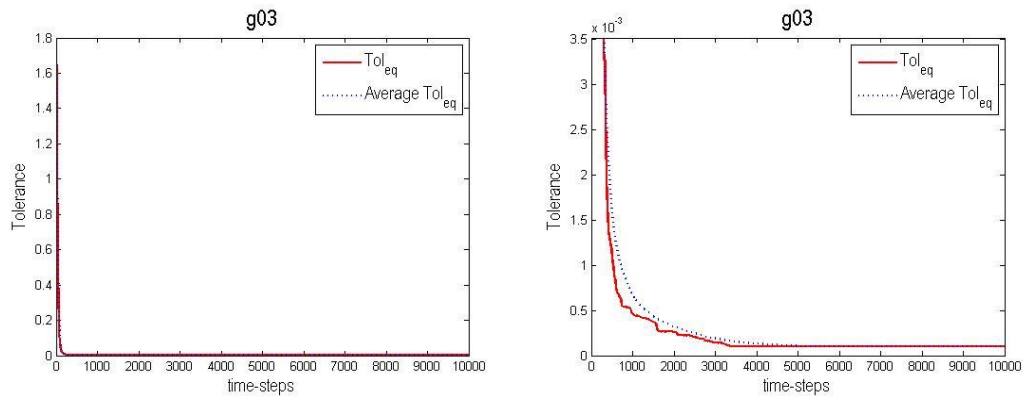

Figure 2: Pseudo-adaptive tolerance for equality constraint violations in problem g03. The average is among the 25 runs. The figure on the right is just a zoom of the one on the left.

The results obtained with the proposed pseudo-adaptive penalization are compared to those obtained by two other PSO algorithms in the literature for reference. Results obtained here are better than those reported by Toscano Pulido et al. [16] for problems g02, g05, g07, g09, g11 and g13, while they are worse for problem g10. They are competitive for the other problems, namely g01, g03, g04, g06, g08, and g12. It is fair to note that Toscano Pulido et al.'s experiments used a smaller number of FEs. Also note that the optimum in Table 2 corresponds to a tolerance for equality constraint violations of $10^{-4}$, while Toscano Pulido et al. used a tolerance of $10^{-3}$. Hence smaller results than the actual minimum feasible solution can be observed in the table.

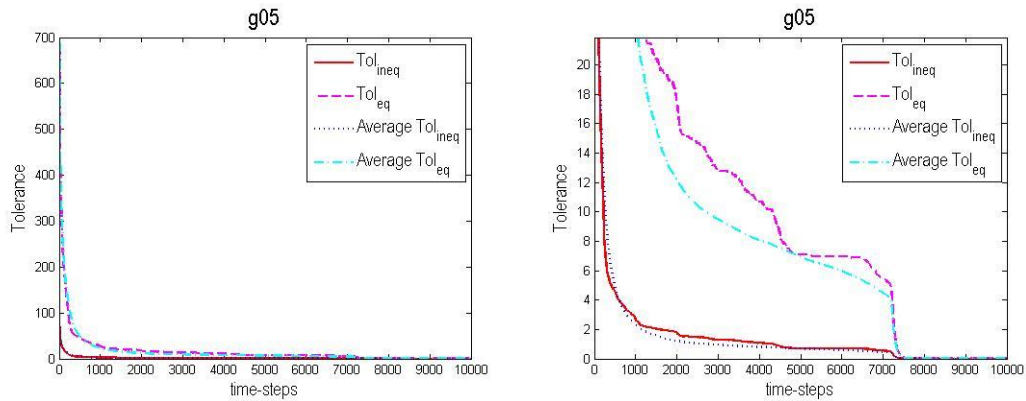

Figure 3: Pseudo-adaptive tolerance for inequality and equality constraint violations in problem g05. The average is among the 25 runs. The figure on the right is just a zoom of the one on the left.

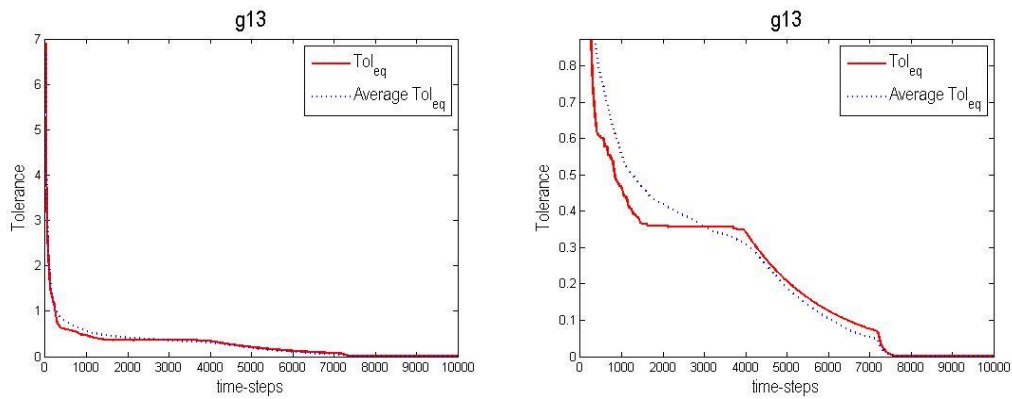

Figure 4: Pseudo-adaptive tolerance for equality constraint violations in problem g13. The average is among the 25 runs. The figure on the right is just a zoom of the one on the left.

Results obtained here are better than those reported by Muñoz Zavala et al. [17] for problems g02 and g13, while they are worse for problems g05, g07, g09 and g10. They are competitive for the remaining problems (g01, g03, g04, g06, g08, g11 and g12). The number of FEs carried out in [17] are reported to be smaller (it is not clear whether they count or not the FEs performed by their mutation operators).

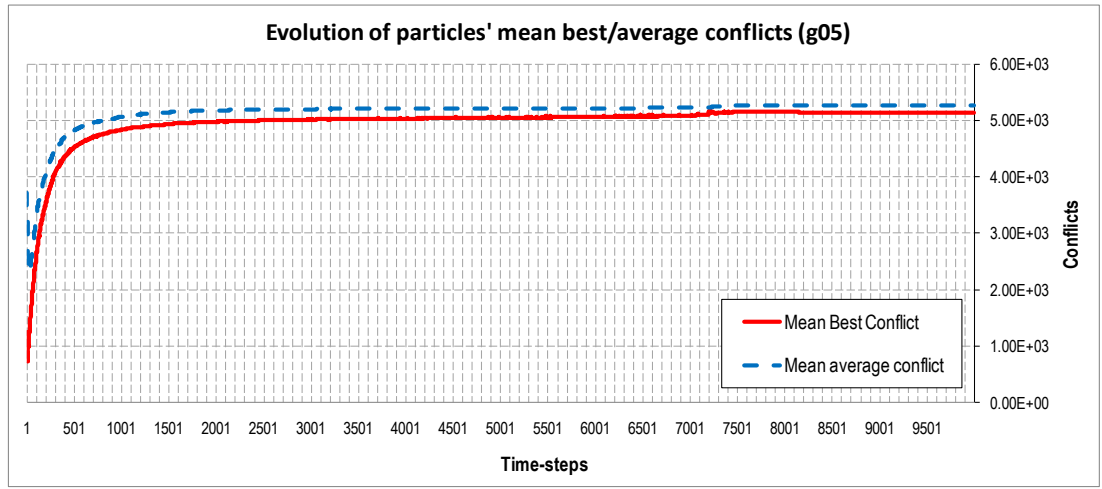

Figure 5: Evolution of the mean best and average conflict values among 25 runs for problem g05. Recall that this is a minimization problem. The shape of the curve is due to the tolerance relaxations.

There are several aspects of the proposed pseudo-adaptive penalization scheme which remain to be studied more thoroughly. Namely, the study of the possible improvement of simply normalizing the constraint violations so that more sensitive constraints do not overtake the search; the initial FR (20–25% here); the percentage of feasible PBESTs that triggers the tolerance update (80% here); the setting of $ktol_{min}$; the calculation of the pseudo-adaptive coefficient $ktol$ (see Equation (13)); and the scheme used to force an update if the desired percentage of feasible PBESTs is not being achieved (see Equation (15)). Preliminary tests showed that the initial FR is not critical within a range such as [1–100%], as the greater the initial FR the more updates of the tolerances are carried out early in the search, as achieving 80% of feasible PBESTs is not difficult. It seems that the results are more sensitive to the setting of $ktol_{min}$; the pseudo-adaptive scheme for the update of $ktol$; and the percentage of feasible PBESTs required to triggering the updates. A more extensive and systematic study of these aspects still needs to be carried out.

# References

[1] J. Kennedy and R. C. Eberhart, "Particle Swarm Optimization," in *Proceedings of the IEEE International Conference on Neural Networks*, Piscataway, 1995.

[2] J. Kennedy and R. C. Eberhart, Swarm Intelligence, Morgan Kaufmann Publishers, 2001.

[3] J. Kennedy and R. C. Eberhart, "A discrete binary version of the particle swarm algorithm," in *Proceedings of the Conference on Systems, Man, and Cybernetics*, Piscataway, 1997.

[4] C. K. Mohan and B. Al-Kazemi, "Discrete particle swarm optimization," in *Proceedings of the Workshop on Particle Swarm Optimization*, Indianapolis, 2001.

[5] M. Clerc, "Discrete Particle Swarm Optimization, illustrated by the Travelling Salesman Problem," in *New Optimization Techniques in Engineering*, Springer-Verlag, 2004, pp. 219-238.

[6] E. Ozcan and C. K. Mohan, "Analisis of a simple particle swarm optimization system," in *Intelligent Engineering Systems Through Artificial Neural Networks: Volume 8*, ASME books, 1998, pp. 253-258.

[7] E. Ozcan and C. K. Mohan, "Particle Swarm Optimization: Surfing the Waves," in *Proceedings of the IEEE Congress on Evolutionary Computation*, Washington, DC, 1999.

[8] F. van den Bergh, An Analysis of Particle Swarm Optimizers, Pretoria: (Ph.D. Thesis) University of Pretoria, 2001.

[9] M. Clerc and J. Kennedy, "The Particle Swarm—Explosion, Stability, and Convergence in a Multidimensional Complex Space," *IEEE Transactions on Evolutionary Computation, Vol. 6, No. 1,* pp. 58-73, 2002.

[10] I. C. Trelea, "The particle swarm optimization algorithm: convergence analysis and parameter selection," *Information Processing Letters 85,* pp. 317-325, 2003.

[11] M. S. Innocente, Development and testing of a Particle Swarm Optimizer to handle hard unconstrained and constrained problems (Ph.D. Thesis), Swansea: Swansea University, 2010.

[12] A. P. Engelbrecht, Fundamentals of Computational Swarm Intelligence, John Wiley & Sons Ltd, 2005.

[13] M. Clerc, Particle Swarm Optimization, Iste, 2006.

[14] K. E. Parsopoulos and M. N. Vrahatis, "Particle Swarm Optimization Method for Constrained Optimization Problems," in *Proceedings of the Euro-International Symposium on Computational Intelligence*, 2002.

[15] C. A. Coello Coello, "Use of a self-adaptive penalty approach for engineering optimization problems," *Computers in Industry,* vol. Vol. 41, pp. 113-127, 2000.

[16] G. Toscano Pulido and C. A. Coello Coello, "A Constraint-Handling Mechanism for Particle Swarm Optimization," in *Proceedings of the IEEE Congress on Evolutionary Computation*, Portland, 2004.

[17] Á. E. Muñoz Zavala, A. Hernández Aguirre and E. R. Villa Diharce, "Constrained Optimization via Particle Evolutionary Swarm Optimization Algorithm (PESO)," in *Proceedings of the 2005 Genetic and Evolutionary Computation Conference (GECCO'05)*, Washington, DC, 2005.

[18] C. W. Reynolds, "Flocks, Herds, and Schools: A Distributed Behavioral Model," *Computer Graphics,* vol. Vol. 21, no. No. 4, pp. 25-34, 1987.

[19] F. Heppner and U. Grenander, "A stochastic nonlinear model for coordinated bird flocks," in *The Ubiquity of Chaos*, Washington, DC, AAAS Publications, 1990, pp. 233-238.